  \documentclass[10pt,twocolumn,letterpaper]{article}

  \usepackage{cvpr}
  \usepackage{times}
  \usepackage{epsfig}
  \usepackage{graphicx}
  \usepackage{amsmath}
  \usepackage{amssymb}
  \usepackage[misc]{ifsym}
  \usepackage{pdfpages}
  \usepackage{graphicx}  

  \usepackage{tabularx, booktabs, multirow, amssymb}
  \newcolumntype{L}[1]{>{\raggedright\let\newline\\\arraybackslash\hspace{0pt}}m{#1}}
  \newcolumntype{C}[1]{>{\centering\let\newline\\\arraybackslash\hspace{0pt}}m{#1}}
  \newcolumntype{R}[1]{>{\raggedleft\let\newline\\\arraybackslash\hspace{0pt}}m{#1}}

  \usepackage[pagebackref=true,breaklinks=true,letterpaper=true,colorlinks,bookmarks=false]{hyperref}

  \cvprfinalcopy 

  \ifcvprfinal\fi
  
\begin{document}

  \title{DeepVoting: A Robust and Explainable Deep Network \\ for Semantic Part Detection under Partial Occlusion}

  \author{Zhishuai Zhang$^{1*}$~~~~~Cihang Xie$^{1*}$~~~~~Jianyu Wang$^{2}$\thanks{The first three authors contributed equally to this work.}~~~~~Lingxi Xie$^{1(\textrm{\Letter})}$~~~~~Alan L. Yuille$^{1}$\\
  Johns Hopkins University$^{1}$~~~Baidu Research USA$^{2}$\\
  {\tt\small \{zhshuai.zhang,~cihangxie306,~wjyouch,~198808xc,~alan.l.yuille\}@gmail.com}\\
  }

  \maketitle
  \thispagestyle{empty}
  \begin{abstract}
  In this paper, we study the task of detecting semantic parts of an object, e.g., a wheel of a car, under partial occlusion. We propose that all models should be trained without seeing occlusions while being able to transfer the learned knowledge to deal with occlusions. This setting alleviates the difficulty in collecting an exponentially large dataset to cover occlusion patterns and is more essential. In this scenario, the proposal-based deep networks, like RCNN-series, often produce unsatisfactory results, because both the proposal extraction and classification stages may be confused by the irrelevant occluders. To address this, \cite{wang2017detecting} proposed a voting mechanism that combines multiple local visual cues to detect semantic parts. The semantic parts can still be detected even though some visual cues are missing due to occlusions. However, this method is manually-designed, thus is hard to be optimized in an end-to-end manner.

  In this paper, we present DeepVoting, which incorporates the robustness shown by \cite{wang2017detecting} into a deep network, so that the whole pipeline can be jointly optimized. Specifically, it adds two layers after the intermediate features of a deep network,  e.g., the pool-4 layer of VGGNet. The first layer extracts the evidence of local visual cues, and the second layer performs a voting mechanism by utilizing the spatial relationship between visual cues and semantic parts.
  We also propose an improved version DeepVoting+ by learning visual cues from context outside objects.
  In experiments, DeepVoting achieves significantly better performance than several baseline methods, including Faster-RCNN, for semantic part detection under occlusion. In addition, DeepVoting enjoys explainability as the detection results can be diagnosed via looking up the voting cues.
  \end{abstract}

  \section{Introduction}
  \label{Introduction}

  Deep networks have been successfully applied to a wide range of vision tasks, in particular object detection~\cite{Girshick_2015_Fast,Ren_2015_Faster,redmon2016you,liu2016ssd, zhang2017single}. Recently, object detection is dominated by a family of proposal-based approaches~\cite{Girshick_2015_Fast,Ren_2015_Faster}, which first generates a set of object proposals for an image, followed by a classifier to predict objects' score for each proposal. However, semantic part detection, despite its importance, has been much less studied. A {\em semantic part} is a fraction of an object which has semantic meaning and can be verbally described, such as a {\em wheel} of a {\em car} or a {\em chimney} of a {\em train}. Detecting semantic parts is a human ability, which enables us to recognize or parse an object at a finer scale.

\begin{figure*}[ht!]
  \centering
  \includegraphics[width=\textwidth]{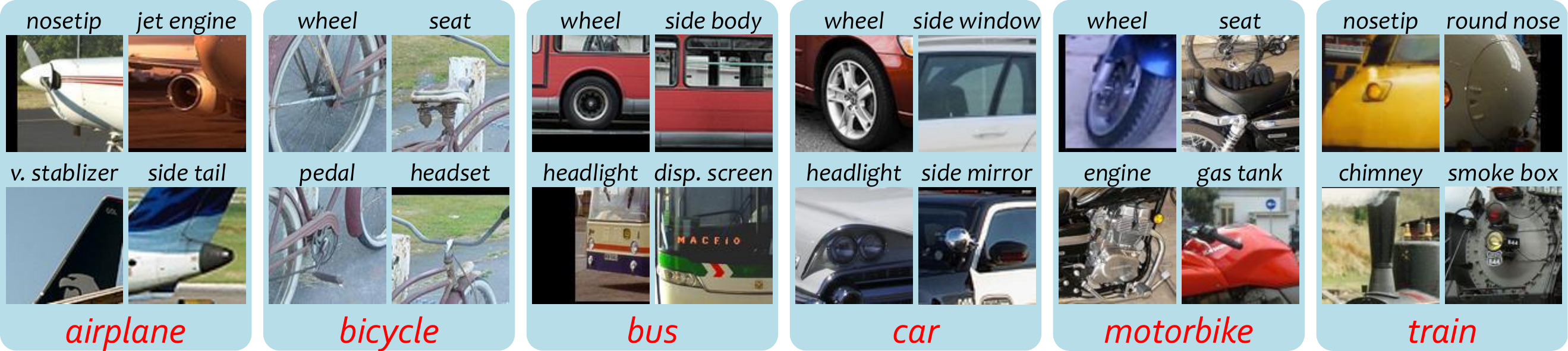}
  \caption{
  Typical semantic parts on six types of rigid objects from the VehicleSemanticPart dataset~\cite{wang2017detecting}. Some semantic parts ({\em e.g.}, {\em wheel}) can appear in different object classes, while some others ({\em e.g.}, {\em chimney}) only appear in one class ({\em train}).
  }
  \label{Fig:SemanticParts}
  \end{figure*}

  In the real world, semantic parts of an object are frequently occluded, which makes detection much harder. In this paper, we investigate semantic part detection especially when these semantic parts are partially or fully occluded. We use the same datasets as in \cite{wang2017detecting}, {\em i.e.}, the VehicleSemanticPart dataset and the VehicleOcclusion dataset. Some typical semantic part examples are shown in Figure \ref{Fig:SemanticParts}. Note that, the VehicleOcclusion dataset is a synthetic occlusion dataset, where the target object is randomly superimposed with two, three or four irrelevant objects (named {\em occluders}) and the occlusion ratios of the target object is constrained. To the best of our knowledge, VehicleOcclusion is the {\em only} public occlusion dataset that provides accurate occlusion annotations of semantic parts like the occlusion ratio and number of occluders. This allows us to evaluate different methods under different occlusion difficulty levels.

  One intuitive solution of dealing with occlusion is to train a model on the dataset that covers different occlusion cases. However, it is extremely difficult yet computationally intractable to collect a dataset that covers occlusion patterns of different numbers, appearances and positions. To overcome this difficulty, we suggest a more essential solution, {\em i.e}, training detectors {\em only} on occlusion-free images, but allowing the learned knowledge ({\em e.g.}, the spatial relationship between semantic parts, {\em etc.}) to be transferred from non-occlusion images to occlusion images. This motivates us to design models that are inherently robust to occlusions. A related work is \cite{wang2017detecting}, which pointed out that proposal-based deep networks are less robust to occlusion, and instead proposed a voting mechanism that accumulates evidences from multiple local visual cues, and locate the semantic parts with the help of geometric constraints ({\em i.e.}, the spatial relationship between the visual cues and the target semantic part). However, this manually-designed framework is broken down into several stages, and thus it is difficult to optimize it in an end-to-end manner. This motivates us to see if the robustness shown in \cite{wang2017detecting} can be incorporated into a deep network which enables end-to-end training naturally.

  To this end, we propose {\bf DeepVoting}, an end-to-end framework for semantic part detection under partial occlusion. Specifically, we add two convolutional layers after the intermediate features of a deep neural network, {\em e.g.}, the neural responses at the {\em pool-4} layer of VGGNet. The first convolutional layer performs template matching and outputs local visual cues named {\em visual concepts}, which were verified to be capable of detecting semantic parts~\cite{Wang_2017_VC_journal}. This layer is followed by a ReLU activation~\cite{Nair_2010_Rectified}, which sets a threshold for filtering the matched patterns, and a dropout layer~\cite{Srivastava_2014_Dropout}, which allows part of evidences to be missing. After that, the second convolution layer is added to perform a voting mechanism by utilizing the spatial relationship between visual cues and semantic parts. The spatial/geometric relations are stored as convolutional weights and visualized as {\em spatial heatmaps}.
  The visual concepts and spatial heatmaps can be learned either on foreground objects only or on whole image with context. We first follow~\cite{wang2017detecting} to train our model on foreground objects only by cropping the object bounding boxes. We further show that visual concepts and spatial heatmaps can also exploit context information by using the whole image to train our model, and we call this improved version {\bf DeepVoting+}.

  We investigate both DeepVoting and DeepVoting+ in our experiments. The first version, in which contexts are excluded, significantly outperforms~\cite{wang2017detecting} with the same setting, arguably because the end-to-end training manner provides a stronger method for joint optimization. The second version, which allows contextual cues to be incorporated, fits the training data better and consequently produces higher detection accuracies. In comparison to the state-of-the-art object detectors such as Faster-RCNN~\cite{Ren_2015_Faster}, DeepVoting enjoys a consistent advantage, and the advantage becomes more significant as the occlusion level goes up. DeepVoting brings two additional benefits apart from being robust to occlusion: (i) DeepVoting enjoys much lower model complexity, {\em i.e.}, the number of parameters is one order of magnitude smaller, and the average testing speed is $2.5\times$ faster; and (ii) DeepVoting provides the possibility to interpret the detection results via looking up the voting cues.


  \section{Related Work}
  \label{RelatedWork}

  Deep convolutional neural networks have been applied successfully to a wide range of computer vision problems, including image recognition~\cite{Krizhevsky_2012_ImageNet,Simonyan_2015_Very,Szegedy_2015_Going,He_2016_Deep}, semantic segmentation~\cite{Long_2015_Fully,Chen_2016_DeepLab,Zheng_2015_Conditional}, object detection~\cite{Girshick_2015_Fast,Ren_2015_Faster,redmon2016you,liu2016ssd,zhang2017single}, {\em etc}. For object detection, one of the most popular pipeline~\cite{Girshick_2015_Fast,Ren_2015_Faster} involved first extracting a number of regions named object proposals~\cite{Alexe_2012_Measuring,Uijlings_2013_Selective,kuo2015deepbox,Ren_2015_Faster}, and then determining if each of them belongs to the target class. Bounding-box regression and non-maximum suppression were attached for post-processing. This framework significantly outperforms the deformable part-based model~\cite{Felzenszwalb_2010_Object} trained on top of a set of handcrafted features~\cite{Dalal_2005_Histograms}.

  There are some works using semantic parts to assist object detection~\cite{chen2014detect,Zhu_2015_DeePM}. Graphical model was used to assemble parts into an object. Also, parts can be used for fine-grained object recognition~\cite{Zhang_2014_Part, zhang2016spda}, be applied as auxiliary cues to understand classification~\cite{Huang_2016_Part}, or be trained for action and attribute classification~\cite{gkioxari2015actions}. Besides, \cite{novotny2016have} investigated the transferability of semantic parts across a large  target  set  of  visually  dissimilar  classes in image understanding.

  Detecting semantic parts under occlusion is an important problem but was less studied before. \cite{wang2017detecting} combined multiple visual concepts via the geometric constraints, {\em i.e.}, the spatial distribution of the visual concepts related to the target semantic parts, to obtain a strong detector. Different from \cite{wang2017detecting}, DeepVoting implements visual concept extraction and the geometric relationships as two layers, and attach them directly to the intermediate outputs of a deep neural network to perform an end-to-end training. This yields much better performance compared to \cite{wang2017detecting}.

  \section{The DeepVoting Framework}
  \label{Framework}

  \begin{figure*}[t!]
  \centering
  \includegraphics[width=\textwidth]{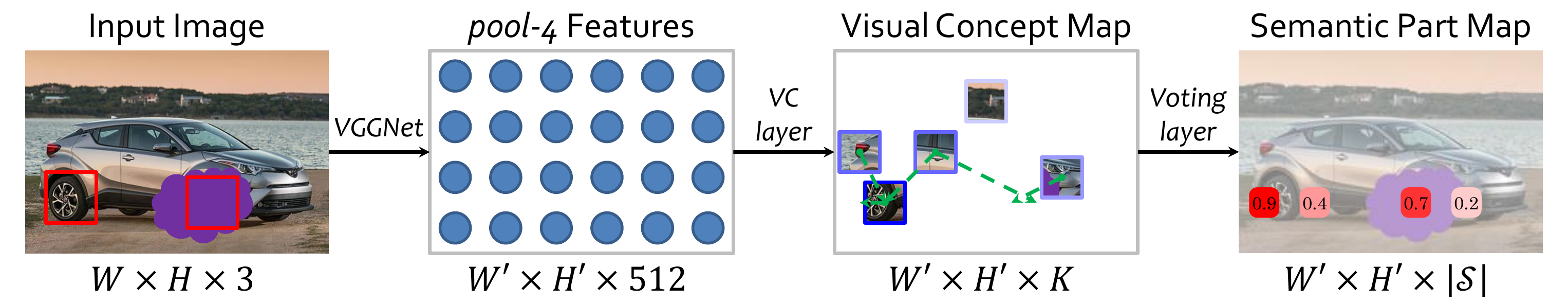}
  \caption{
  The overall framework of DeepVoting (best viewed in color). A {\em car} image with two {\em wheels} (marked by red frames, one of them is occluded) is fed into VGGNet~\cite{Simonyan_2015_Very}, and the intermediate outputs are passed through a visual concept extraction layer and a voting layer. We aggregate local cues from the visual concept map (darker blue indicates more significant cues), consider their spatial relationship to the target semantic part via voting, and obtain a low-resolution map of semantic parts (darker red or a larger number indicates higher confidence). Based on this map, we perform bounding box regression followed by non-maximum suppression to obtain the final results.
  }
  \label{Fig:Framework}
  \end{figure*}

  \subsection{Motivation}
  \label{Framework:Motivation}

  We aim at detecting the semantic parts of an object under occlusion. First of all, we argue that {\em only} occlusion-free images should be used in the training phase. This is because the appearance and position of the occluders can be arbitrary, thus it is almost impossible to cover all of them by a limited training set. It is our goal to design a framework which can transfer the learned knowledge from the occlusion-free domain to the occlusion domain.

  One possible solution is to adapt the state-of-the-art object detection methods, such as Faster-RCNN~\cite{Ren_2015_Faster}, to detect semantic parts. Specifically, the adapted methods first extract a number of proposals for semantic parts and then compute the classification scores for each of them. But, we point out that this strategy may miss some partially or fully occluded semantic parts because of two important factors: (1) occlusion may distract the proposal generation network from extracting good proposals for semantic parts; (2) even with correct proposals, the classifier may still go wrong since the appearance of the occluded semantic parts can be totally different. We verify that these factors indeed downgrade the performance of Faster-RCNN in Section~\ref{Experiments:Occlusion}.

  The voting mechanism~\cite{wang2017detecting} suggests an alternative strategy, which accumulates mid-level visual cues to detect high-level semantic parts. These mid-level cues are called visual concepts~\cite{Wang_2017_VC_journal},  {\em i.e.}, a set of intermediate CNN states which are closely related to semantic parts. A semantic part is supported by multiple visual concepts via the geometric constraints between them. Even if the evidences from some visual concepts are missing due to occlusion, it is still possible to infer the presence of the semantic part via the evidences from the remaining ones. However, it involves too many hyper-parameters and thus is hard to be optimized.

  In this paper, we propose DeepVoting which incorporates the robustness shown by \cite{wang2017detecting} into a deep network. Following \cite{Wang_2017_VC_journal}, the visual concepts are learned when the objects appear at a fixed scale since each neuron on the intermediate layer, {\em e.g.}, the {\em pool-4} layer, has a fixed receptive field size~\cite{Simonyan_2015_Very}. Therefore, we assume that the object scale is approximately the same in both training and testing stages. In the training stage, we used the ground-truth bounding box to resize the object for the DeepVoting, and compute the object-to-image ratio to train a standalone network, ScaleNet~\cite{Qiao_2017_ScaleNet}, for scale prediction (see Section~\ref{Framework:ScaleNet} for details). In the testing stage, the trained ScaleNet was used to predict the resizing ratio, and then we resize the testing image according to the predicted ratio.

  \subsection{Formulation}
  \label{Framework:Formulation}

  Let $\mathbf{I}$ denote an image with a size of $W\times H \time 3$. Following~\cite{wang2017detecting}, we feed this image into a $16$-layer VGGNet~\cite{Simonyan_2015_Very}, and extract the {\em pool-4} features as a set of intermediate neural outputs. Denote the output of the {\em pool-4} layer as $\mathbf{X}$, or a $W'\times H'\times D$ cube, where $W'$ and $H'$ are the down-sampled scales of $W$ and $H$, and ${D}$ is 512 for VGGNet. These features can be considered as $W'\times H'$ high-dimensional vectors, and each of them represents the appearance of a local region. Denote each $D$-dimensional feature vector as $\mathbf{x}_i$ where $i$ is an index at the $W'\times H'$ grid. These feature vectors are $\ell_2$-normalized so that ${\left\|\mathbf{x}_i\right\|_2}={1}$.

  \subsubsection{Visual Concept Extraction}
  \label{Framework:Formulation:Extraction}

  In~\cite{wang2017detecting}, a set of visual concepts ${\mathcal{V}}={\left\{\mathbf{v}_1,\ldots,\mathbf{v}_K\right\}}$ are obtained via K-means clustering, and each visual concept is considered intuitively as a template to capture the mid-level semantics from these intermediate outputs. Specifically, the response of the visual concept $\mathbf{v}_k$ at the {\em pool-4} feature vector $\mathbf{x}_i$ is measured by the $\ell_2$-distance, {\em i.e.}, $\left\|\mathbf{v}_k-\mathbf{x}_i\right\|_2^2$.

  We note that $\mathbf{x}_i$ has unit length, and so ${\left\|\mathbf{v}_k\right\|_2}\approx{1}$ as it is averaged over a set of neighboring $\mathbf{x}_i$'s, so we have ${\left\|\mathbf{v}_k-\mathbf{x}_i\right\|_2^2}\approx{2-2\left\langle\mathbf{v}_k,\mathbf{x}_i\right\rangle}$ where $\left\langle\cdot,\cdot\right\rangle$ is the dot product operator. Then the log-likelihood ratio tests are applied to eliminate negative responses. This is driven by the idea that the presence of a visual concept can provide positive cues for the existence of a semantic part, but the absence of a visual concept shall not give the opposite information.

  Different from \cite{wang2017detecting}, DeepVoting implements this module as a convolutional layer, namely visual concept layer, and attaches it directly after the normalized intermediate outputs of a deep neural network. The kernel size of this convolutional layer is set to be $1\times1$, {\em i.e.}, each $\mathbf{x}_i$ is considered individually. The ReLU activation~\cite{Nair_2010_Rectified} follows to set the negative responses as $0$'s and thus avoids them from providing negative cues. We append a dropout layer~\cite{Srivastava_2014_Dropout} with a drop ratio $0.5$, so that a random subset of the visual concept responses are discarded in the training process.
  This strategy facilitates the model to perform detection robustly using incomplete information and, consequently, improves the testing accuracy when occlusion is present.

  The output of visual concept layer is a map $\mathbf{Y}$ of size $W'\times H'\times\left|\mathcal{V}\right|$, where $\mathcal{V}$ is the set of visual concepts. We set ${\left|\mathcal{V}\right|}={256}$, though a larger set may lead to slightly better performance. Although these visual concepts are trained from scratch rather than obtained from clustering~\cite{Wang_2017_VC_journal}, we show in Section~\ref{Experiments:NonOcclusion:Visualization} that they are also capable of capturing repeatable visual patterns and semantically meaningful.

  \subsubsection{Semantic Part Detection via the Voting Layer}
  \label{Framework:Formulation:Voting}

  After the previous stage, we can find some fired visual concepts, {\em i.e.}, those positions with positive response values. In \cite{wang2017detecting}, the fired visual concepts are determined via log-likelihood ratio tests. These {\em fired} visual concepts are then accumulated together for semantic part detection by considering the spatial constraints between each pair of visual concept and semantic part. It is motivated by the nature that a visual concept can, at least weakly, suggest the existence of a semantic part. For example, as shown in Figure~\ref{Fig:Framework}, in a {\em car} image, finding a {\em headlight} implies that there is a {\em wheel} nearby, and the distance and direction from the {\em headlight} to the {\em wheel} are approximately the same under a fixed scale.

  Different from \cite{wang2017detecting}, DeepVoting implements the spatial constraints between visual concepts and the semantic parts as another convolutional layer, named the voting layer, in which we set the receptive field of each convolutional kernel to be large, {\em e.g.}, $15\times15$, so that a visual concept can vote for the presence of a semantic part at a relatively long distance. This strategy helps particularly when the object is partially occluded, as effective visual cues often emerge outside the occluder and may be far from the target.

  Though the spatial constraints are learned from scratch and only semantic part level supervision is imposed during training, they can still represent the frequency that  visual concepts appear at different relative positions. We refer to each learned convolutional kernel at this layer as a spatial {\em heatmap}, and some of them are visualized in Section~\ref{Experiments:NonOcclusion:Visualization}.

  Denote the output of the voting layer, {\em i.e.}, the semantic part map, as $\mathbf{Z}$. It is a $W'\times H'\times\left|\mathcal{S}\right|$ cube where $\mathcal{S}$ is the set of semantic parts. Each local maximum at the semantic part map corresponds to a region on the image lattice according to their receptive filed. To generate a bounding box for semantic part detection, we first set an anchor box, sized $100\times100$ and centered at this region, and then learn the spatial rescaling and translation to regress the anchor box (following the same regression procedure in \cite{Girshick_2015_Fast}) from the training data. The anchor size $100\times100$ is the average semantic part scale over the entire training dataset~\cite{wang2017detecting}.

  \begin{figure*}[t!]
  \centering
  \includegraphics[width=.233\textwidth]{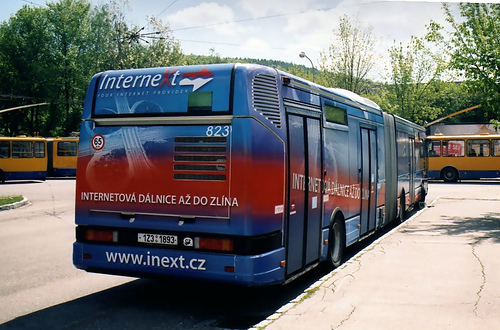} \hfill
  \includegraphics[width=.233\textwidth]{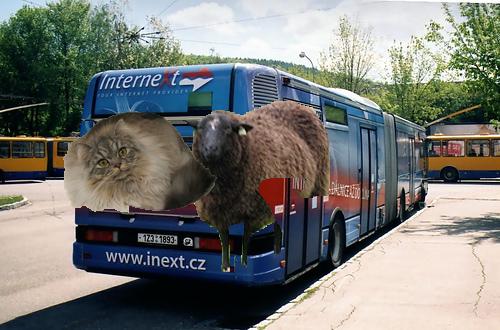} \hfill
  \includegraphics[width=.233\textwidth]{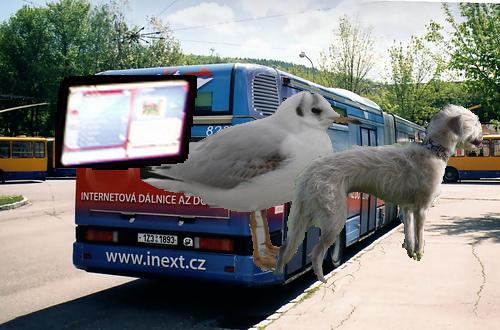} \hfill
  \includegraphics[width=.233\textwidth]{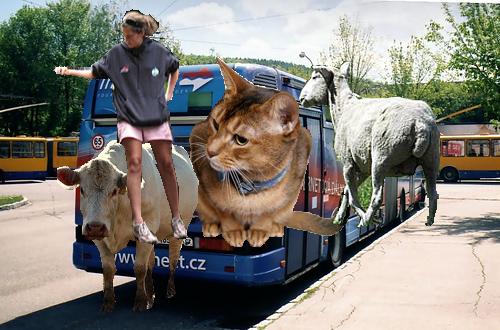}
  \caption{
  Examples of images in VehicleSemanticPart dataset and VehicleOcclusion dataset. The first is the original occlusion-free image from VehicleSemanticPart dataset. The second, third and forth image (in row-major order) are from VehicleOcclusion dataset. There are $2$, $3$ and $4$ occluders, and the occluded ratio of object, computed by pixels, is $0.2$--$0.4$, $0.4$--$0.6$ and $0.6$--$0.8$, respectively.
  }
  \label{Fig:Occlusion}
  \end{figure*}

  \subsection{Training and Testing}
  \label{Framework:Details}

  We train the network on an occlusion-free image corpus. This helps us obtain clear relationship between the visual concepts and the semantic parts. We discard the background region by cropping the object according to the ground-truth bounding box, to be consistent with~\cite{wang2017detecting}. Then, we rescale the cropped image so that the object short edge has $224$ pixels, which is motivated by~\cite{Wang_2017_VC_journal} to capture the visual concepts at a fixed scale. The image is fed into the $16$-layer VGGNet, and we get the feature vectors at the {\em pool-4} layer.

  These feature vectors are normalized and passed through two layers for visual concept extraction and voting. We compare the output semantic part map $\mathbf{Z}$ with the ground-truth annotation $\mathbf{L}$ by computing dice coefficient between prediction and ground-truth~\cite{Milletari_2016_V}. To generate the ground-truth, we find the nearest grid point at the $W'\times H'$ grid (down-sampled from the original image by the factor of $16$) based on the center pixel of each annotated semantic part, and set the labels of these positions as $1$ and others as $0$. Then we apply Gaussian filtering on the binary ground-truth annotation, to generate the smoothed ground-truth annotation $\mathbf{L}$. The label cube $\mathbf{L}$ is also of size $W'\times H'\times\left|\mathcal{S}\right|$. The similarity between $\mathbf{Z}$ and $\mathbf{L}$ is defined as:
  \begin{equation}
  \label{Eqn:ProbPositive}
  {\mathcal{D}\!\left(\mathbf{Z},\mathbf{L}\right)}={\frac{1}{\left|\mathcal{S}\right|}{\sum_{s=1}^{\left|\mathcal{S}\right|}}\frac{2\times{\sum_{w=1,h=1}^{W',H'}}z_{w,h,s}\times l_{w,h,s}}{{\sum_{w=1,h=1}^{W',H'}}\left(z_{w,h,s}^2+l_{w,h,s}^2\right)}},
  \end{equation}
  It is straightforward to compute the gradients based on the loss function ${\mathcal{L}\!\left(\mathbf{Z},\mathbf{L}\right)}={1-\mathcal{D}\!\left(\mathbf{Z},\mathbf{L}\right)}$.

  On the testing stage, we first use ScaleNet (see Section~\ref{Framework:ScaleNet}) to obtain the object scale. Then, we rescale the image so that the short edge of the object roughly contains $224$ pixels. We do not crop the object because we do not know its location. Then, the image is passed through the VGGNet followed by both visual concept extraction and voting layers, and finally we apply the spatial rescaling and translation to the anchor box ($100\times100$) towards more accurate localization. A standard non-maximum suppression is performed to finalize the detection results.

  DeepVoting is trained on the images cropped with respect to the object bounding boxes to be consistent with~\cite{wang2017detecting}. Moreover, visual concepts and spatial heatmaps can also exploit context outside object bounding boxes. To verify this, we train an improved version, named DeepVoting+, without cropping the bounding boxes. We also resize the image so that the object short edge contains 224 pixels in the training stage, and the testing stage is the same as DeepVoting. Experiments show that DeepVoting+ achieves better performance than DeepVoting.

  \subsection{The Scale Prediction Network}
  \label{Framework:ScaleNet}

  The above framework is based on an important assumption, that the objects appear in approximately the same scale. This is due to two reasons. First, as shown in~\cite{Wang_2017_VC_journal}, the visual concepts emerge when the object is rescaled to the same scale, {\em i.e.}, the short edge of the object bounding box contains $224$ pixels. Second, we expect the voting layer to learn fixed spatial offsets which relate a visual concept to a semantic part. As an example, the heatmap delivers the knowledge that in the side view of a {\em car}, the {\em headlight} often appears at the upperleft direction of a {\em wheel}, and the spatial offset on $x$ and $y$ axes are about $64$ and $48$ pixels ($4$ and $3$ at the {\em pool-4} grid), respectively. Such information is not scale-invariant.

  To deal with these issues, we introduce an individual network, namely the ScaleNet~\cite{Qiao_2017_ScaleNet}, to predict the object scale in each image. The main idea is to feed an input image to a $16$-layer VGGNet for a regression task (the {\em fc-8} layer is replaced by a $1$-dimensional output), and the label is the ground-truth object size. Each input image is rescaled, so that the long edge contains $224$ pixels. It is placed at the center of an $224\times224$ square and the remaining pixels are filled up with the averaged intensity. During the training, we consider the short edge of the object, and ask the deep network to predict the ratio of the object short edge to the image long edge ($224$ pixels). In the test phase, an image is prepared and fed into the network in the same flowchart, and the predicted ratio is used to normalize the object to the desired size, {\em i.e.}, its short edge contains $224$ pixels. We show in Section~\ref{Experiments:NonOcclusion:ScalePrediction} that this method works very well.

  \subsection{Discussions and Relationship to Other Works}
  \label{Framework:Discussions}

  The overall framework of DeepVoting is quite different from the conventional proposal-based detection methods, such as Faster-RCNN~\cite{Ren_2015_Faster}. This is mainly due to the problem setting, {\em i.e.}, when the occlusion is present, the accuracy of both proposal and classification networks becomes lower. However, DeepVoting is able to infer the occluded semantic parts via accumulating those non-occluded visual cues. We show more comparative experiments in Section~\ref{Experiments:Occlusion}.

  We decompose semantic part detection into two steps, {\em i.e.}, central pixel detection and bounding box regression. The first step is performed like semantic segmentation~\cite{Long_2015_Fully} in a very low-resolution setting (down-sampled from the original image by the factor of $16$). We also borrow the idea from segmentation~\cite{Milletari_2016_V}, which uses a loss function related to the dice coefficient in optimization. As the semantic part is often much smaller compared to the entire image, this strategy alleviates the bias of data imbalance, {\em i.e.}, the model is more likely to predict each pixel as background as it appears dominantly in the training data.

  \section{Experiments}
  \label{Experiments}

  \subsection{Dataset and Baseline}
  \label{Experiments:Settings}

  \newcommand{\colwidth}{1.6cm}
  \newcommand{\colwidthA}{0.65cm}
  \begin{table*}[!ht]
  \centering
  {\setlength{\tabcolsep}{0.08cm} 
  \begin{tabular}{|L{\colwidth}||R{\colwidthA}|R{\colwidthA}|R{\colwidthA}|R{\colwidthA}|R{\colwidthA}|R{\colwidthA}||R{\colwidthA}|R{\colwidthA}|R{\colwidthA}|R{\colwidthA}||R{\colwidthA}|R{\colwidthA}|R{\colwidthA}|R{\colwidthA}||R{\colwidthA}|R{\colwidthA}|R{\colwidthA}|R{\colwidthA}|}
  \hline
  \multirow{2}{*}{} & \multicolumn{6}{c||}{No Occlusions} & \multicolumn{4}{c||}{L1} &
  \multicolumn{4}{c||}{L2} & \multicolumn{4}{c|}{L3}                                      \\
  \cline{2-19}
  {Category}          & \multicolumn{1}{c|}{\bf KVC}        & \multicolumn{1}{c|}{\bf DVC}  & \multicolumn{1}{c|}{\bf VT}
  & \multicolumn{1}{c|}{\bf FR}         & \multicolumn{1}{c|}{\bf DV}  & \multicolumn{1}{c||}{\bf DV+}
  & \multicolumn{1}{c|}{\bf VT} & \multicolumn{1}{c|}{\bf FR} & \multicolumn{1}{c|}{\bf DV}  & \multicolumn{1}{c||}{\bf DV+}
  & \multicolumn{1}{c|}{\bf VT} & \multicolumn{1}{c|}{\bf FR} & \multicolumn{1}{c|}{\bf DV}  & \multicolumn{1}{c||}{\bf DV+}
  & \multicolumn{1}{c|}{\bf VT} & \multicolumn{1}{c|}{\bf FR} & \multicolumn{1}{c|}{\bf DV}   & \multicolumn{1}{c|}{\bf DV+}       \\
  \hline\hline
  {\em airplane}  & 15.8                         & 26.6                         & 30.6                        & 56.9                        & 59.0                        & \textbf{60.2} & 23.2                        & 35.4                        & \textbf{40.6}               & \textbf{40.6} & 19.3                        & 27.0                        & 31.4                        & \textbf{32.3} & 15.1                        & 20.1                        & \textbf{25.9}               & 25.4          \\ \hline
  {\em bicycle}   & 58.0                         & 52.3                         & 77.8                        & 90.6               & 89.8                        & \textbf{90.8}          & 71.7                        & 77.0                        & 83.5                        & \textbf{85.2} & 66.3                        & 62.0                        & 78.7                        & \textbf{79.6} & 54.3                        & 41.1                        & \textbf{63.0}               & 62.5          \\ \hline
  {\em bus}       & 23.8                         & 25.1                         & 58.1                        & \textbf{86.3}               & 78.4                        & 81.3          & 31.3                        & 55.5                        & 56.9                        & \textbf{65.8} & 19.3                        & 40.1                        & 44.1                        & \textbf{54.6} & 9.5                         & 25.8                        & 30.8                        & \textbf{40.5} \\ \hline
  {\em car}       & 25.2                         & 36.5                         & 63.4                        & 83.9                        & 80.4                        & \textbf{80.6} & 35.9                        & 48.8                        & 56.1                        & \textbf{57.3} & 23.6                        & 30.9                        & 40.0                        & \textbf{41.7} & 13.8                        & 19.8                        & 27.3                        & \textbf{29.4} \\ \hline
  {\em motorbike} & 32.7                         & 29.2                         & 53.4                        & 63.7                        & 65.2                        & \textbf{69.7} & 44.1                        & 42.2                        & 51.7                        & \textbf{55.5} & 34.7                        & 32.4                        & 41.4                        & \textbf{43.4} & 24.1                        & 20.1                        & 29.4                        & \textbf{31.2} \\ \hline
  {\em train}     & 12.3                         & 12.8                         & 35.5                        & 59.9                        & 59.4                        & \textbf{61.2} & 21.7                        & 30.6                        & 33.6                        & \textbf{43.7} & 8.4                         & 17.7                        & 19.8                        & \textbf{29.8} & 3.7                         & 10.9                        & 13.3                        & \textbf{22.2} \\ \hline
  \textbf{mean}   & 28.0                         & 30.4                         & 53.1                        & 73.6                        & 72.0                        & \textbf{74.0} & 38.0                        & 48.3                        & 53.7                        & \textbf{58.0} & 28.6                        & 35.0                        & 42.6                        & \textbf{46.9} & 20.1                        & 23.0                        & 31.6                        & \textbf{35.2} \\ \hline
  \end{tabular}}
  \caption{
  Left 6 columns: Comparison of detection accuracy (mean AP, $\%$) of {\bf KVC}, {\bf DVC}, {\bf VT}, {\bf FR}, {\bf DV} and {\bf DV+}  without occlusion. Right 12 columns: Comparison of detection accuracy (mean AP, $\%$) of {\bf VT}, {\bf FR}, {\bf DV} and {\bf DV+}   when the object is occluded at three different levels. Note that {\bf DV+} is DeepVoting trained with context outside object bounding boxes. See the texts for details.
  }
  \label{Tab:NoOcclusion}
  \label{Tab:Occlusion}
  \end{table*}

  We use the  VehicleSemanticPart dataset and the VehicleOcclusion dataset~\cite{wang2017detecting} for evaluation. The VehicleSemanticPart dataset contains $4549$ training images and $4507$ testing images covering six types of vehicles, {\em i.e.}, {\em airplane}, {\em bicycle}, {\em bus}, {\em car}, {\em motorbike} and {\em train}. In total, $133$ semantic parts are annotated. For each test image in VehicleSemanticPart dataset, some randomly-positioned occluders (irrelevant to the target object) are placed onto the target object, and make sure that the occlusion ratio of the target object is constrained. Figure \ref{Fig:Occlusion} shows several examples with different occlusion levels.

  We train six models, one for each object class. All the models are trained on an occlusion-free dataset, but evaluated on either non-occluded images, or the images with different levels of occlusions added. In the later case, we vary the difficulty level by occluding different fractions of the object. We evaluate all the competitors following a popular criterion~\cite{Everingham_2010_PASCAL}, which computes the mean average precision (mAP) based on the list of detected semantic parts. A detected box is considered to be true-positive if and only if its IoU rate with a ground-truth box is not lower than $0.5$. Each semantic part is evaluated individually, and the mAP of each object class is the average mAP over all the semantic parts.

  DeepVoting and DeepVoting+ (denoted by {\bf DV} and {\bf DV+}, respectively, Section~\ref{Framework:Details}) are compared with four baselines:
  \begin{itemize}
  \item {\bf KVC}: These visual concepts are clustered from a set of {\em pool-4} features using $K$-Means~\cite{Wang_2017_VC_journal}. The ScaleNet (detailed in Section~\ref{Framework:ScaleNet}) is used to tackle scale issue and the extracted visual concepts are directly used to detect the semantic parts.
  \item {\bf DVC}: These visual concepts are obtained from DeepVoting, {\em i.e.}, the weights of the visual concept extraction layer. The ScaleNet (detailed in Section~\ref{Framework:ScaleNet}) is used to tackle scale issue and the extracted visual concepts are directly used to detect the semantic parts.
  \item {\bf VT}: The voting method first finds fired visual concepts via log-likelihood ratio tests, and then utilizes spatial constraints to combine these local visual cues.
  \item {\bf FR}:   We train models for each category independently. Each semantic part of a category is considered as a separate class during training, {\em i.e.}, for each category, we train a model with $\left|\mathcal{S}\right|+1$ classes, corresponding to $\left|\mathcal{S}\right|$ semantic parts and the background. Different from other baselines, Faster-RCNN here is trained on full images, {i.e.}, object cropping is not required. This enables Faster-RCNN to use context for semantic parts detection and handle scale issue naturally since semantic parts with various scales are used in training.
  \end{itemize}

  \subsection{Semantic Part Detection without Occlusion}
  \label{Experiments:NonOcclusion}

  As a simplified task, we evaluate our algorithm in detecting semantic parts on non-occluded objects. This is also a baseline for later comparison. In the left six columns of Table~\ref{Tab:NoOcclusion}, we list the detection accuracy produced by different methods. The average detection accuracies by both voting and DeepVoting are significantly higher than using single visual concept for detection, regardless whether the visual concepts are obtained from $K$-Means clustering or DeepVoting. This indicates the advantage of the approaches which aggregates multiple visual cues for detection. Meanwhile, DeepVoting is much better than voting due to the better scale prediction and the end-to-end training manner. Even the right scale is provided for voting (oracle scale results in \cite{wang2017detecting}), DeepVoting still beat it by more than $20\%$ in terms of averaged mAP over 6 objects, which indicates the benefit brought by the joint optimization of both weights for visual concept extraction layer and voting layer.

  On the other hand, DeepVoting produces slightly lower detection accuracy compared to Faster-RCNN. We argue that Faster-RCNN benefits from the context outside object bounding boxes, as we can see, if we improve DeepVoting by adding context during the training ({\em i.e.} DeepVoting+), Faster-RCNN will be less competitive compared with our method. Meanwhile, DeepVoting enjoys lower computational overheads, {\em i.e.}, it runs $2.5\times$ faster.

  \newcommand{\colwidthDiagnosis}{0.8cm}
  \begin{table*}[!t]
  \centering
  \begin{tabular}{|l||C{\colwidthDiagnosis}|C{\colwidthDiagnosis}|C{\colwidthDiagnosis}|C{\colwidthDiagnosis}||C{\colwidthDiagnosis}|C{\colwidthDiagnosis}|C{\colwidthDiagnosis}||C{\colwidthDiagnosis}|C{\colwidthDiagnosis}|C{\colwidthDiagnosis}|}
  \hline
  \multirow{2}{*}{} & \multicolumn{4}{c||}{{\bf Recall} at Different Levels}
  & \multicolumn{3}{c||}{{\bf mAP} w/ Addt'l Prop.}
  & \multicolumn{3}{c| }{{\bf mAP} by DeepVoting+}                             \\
  \cline{2-11}
  {Category}          & \multicolumn{1}{c|}{L0}             & \multicolumn{1}{c|}{L1}
  & \multicolumn{1}{c|}{L2}             & \multicolumn{1}{c||}{L3}
  & \multicolumn{1}{c|}{L1}
  & \multicolumn{1}{c|}{L2}             & \multicolumn{1}{c||}{L3}
  & \multicolumn{1}{c|}{L1}
  & \multicolumn{1}{c|}{L2}             & \multicolumn{1}{c|}{L3}             \\
  \hline\hline
  {\em airplane}  & $99.3 $ & $98.1 $ & $97.4 $ & $96.7 $
  & $36.2 $          & $27.7 $          & $20.7 $
  & $\mathbf{40.6 }$ & $\mathbf{32.3 }$ & $\mathbf{25.4 }$                    \\
  \hline
  {\em bicycle}   & $99.5 $ & $99.0 $ & $98.0 $ & $96.5 $
  & $77.9 $          & $64.0 $          & $44.7 $
  & $\mathbf{85.2 }$ & $\mathbf{79.6 }$ & $\mathbf{62.5 }$                    \\
  \hline
  {\em bus}       & $99.8 $ & $96.3 $ & $93.8 $ & $91.5 $
  & $57.1 $ & $42.4 $ & $28.3 $
  & $\mathbf{65.8 } $          & $\mathbf{54.6 }$          & $\mathbf{40.5 }$                    \\
  \hline
  {\em car}       & $99.8 $ & $96.0 $ & $94.4 $ & $92.7 $
  & $48.2 $          & $30.2 $          & $19.4 $
  & $\mathbf{57.3 }$ & $\mathbf{41.7 }$ & $\mathbf{29.4 }$                    \\
  \hline
  {\em motorbike} & $99.0 $ & $96.5 $ & $95.7 $ & $93.3 $
  & $43.6 $          & $33.1 $          & $21.3 $
  & $\mathbf{55.5 }$ & $\mathbf{43.4 }$ & $\mathbf{31.2 }$                    \\
  \hline
  {\em train}     & $98.3 $ & $93.5 $ & $90.6 $ & $85.6 $
  & $32.0 $          & $19.4 $ & $11.3 $
  & $\mathbf{43.7 }$ & $\mathbf{29.8 } $          & $\mathbf{22.2 }$                    \\
  \hline
  {\bf mean}      & $99.3 $ & $96.6 $ & $95.0 $ & $92.7 $
  & $49.2 $          & $36.1 $          & $24.2 $
  & $\mathbf{58.0 }$ & $\mathbf{46.9 }$ & $\mathbf{35.2 }$                    \\
  \hline
  \end{tabular}
  \caption{
  Left $4$ columns: the recall rates ($\%$) of the proposal network at different occlusion levels. Middle $3$ and right $3$ columns: detection mAPs ($\%$) of Faster-RCNN (ground-truth bounding boxes are added as additional proposals) and DeepVoting+ at different occlusion levels.
  }
  \label{Tab:Diagnosis}
  \end{table*}

  \begin{figure}[t!]
  \centering
  \includegraphics[width=0.85\linewidth]{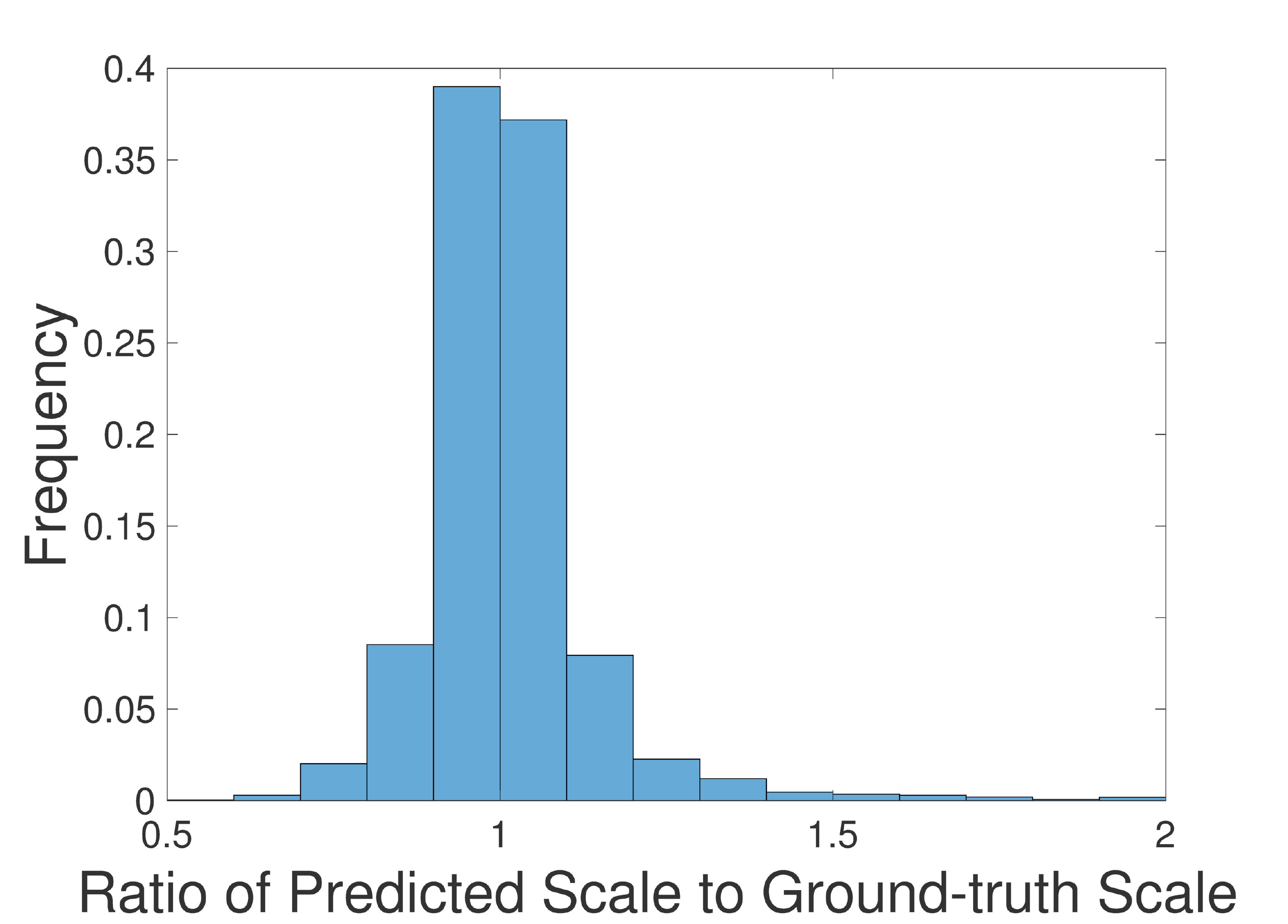}
  \caption{
  The distribution of the ratio of the predicted scale to the actual scale.
  }
  \label{Fig:ScalePrediction}
  \end{figure}

  \subsubsection{Scale Prediction Accuracy}
  \label{Experiments:NonOcclusion:ScalePrediction}

  We investigate the accuracy of ScaleNet, which is essential for scale normalization. For each testing image, we compute the ratio of the predicted object scale to the actual scale, and plot the contribution of this ratio over the entire testing set in Figure~\ref{Fig:ScalePrediction}. One can see that in more than $75\%$ cases, the relative error of the predicted scale does not exceed $10\%$. Actually, these prediction results are accurate enough for DeepVoting. Even if ground-truth scale is provided and we rescale the images accordingly, the detection accuracy is slightly improved from $72.0\%$ to $74.5\%$.

  \begin{figure*}[t!]
  \centering
  \includegraphics[width=\textwidth]{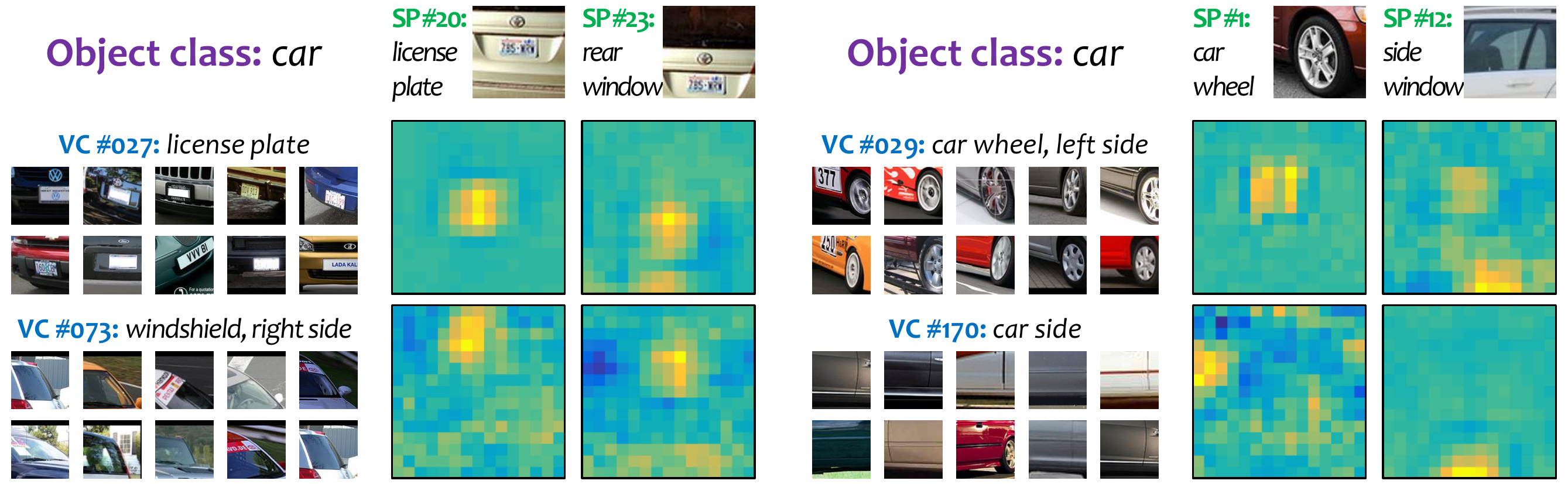}
  \caption{
  Visualization of visual concepts and spatial heatmaps (best viewed in color). For each visual concept, we show $10$ patches with the highest responses. Each spatial heatmap illustrates the cues to detect a semantic part, in which yellow, cyan and dark blue indicate positive, zero and negative cues, respectively. For example, {\tt VC \#073} ({\em windshield}) often appears above {\tt SP \#20} ({\em license plate}), and {\tt VC \#170} ({\em car side bottom}) often appears below {\tt SP \#12} ({\em side window}).
  }
  \label{Fig:Visualization}
  \end{figure*}

  \subsection{Semantic Part Detection under Occlusion}
  \label{Experiments:Occlusion}
  We further detect semantic parts when the object is occluded in three different levels. Since the baselines {\bf KVC} and {\bf DVC} perform much worse than other methods even when occlusion is not present, we ignore these two methods when performing semantic part detection under occlusion. In the first level ({\em i.e.} {\bf L1}), we place $2$ occluders on each object, and the occluded ratio $r$ of the object, computed by pixels, satisfying ${0.2}\leqslant{r}<{0.4}$. For {\bf L2} and {\bf L3}, we have $3$ and $4$ occluders, and ${0.4}\leqslant{r}<{0.6}$ and ${0.6}\leqslant{r}<{0.8}$, respectively (see Figure~\ref{Fig:Occlusion} for examples). The original occlusion-free testing set is denoted as {\bf L0}. The detection results are summarized in Table~\ref{Tab:Occlusion}. One can see that DeepVoting outperforms the voting and the Faster-RCNN significantly in these cases. For the Faster-RCNN, the accuracy gain increases as the occlusion level goes up, suggesting the advantage of DeepVoting in detecting occluded semantic parts. As a side evidence, we investigate the impact of the size of spatial heatmap (the kernel of the voting layer). At the heaviest occlusion level, when we shrink the default $15\times15$ to $11\times11$, the mean detection accuracy drops from $31.6\%$ to $30.6\%$, suggesting the usefulness of long-distance voting in detecting occluded semantic parts. When the kernel size is increased to $19\times19$, the accuracy is slightly improved to $31.8\%$. Therefore, we keep the kernel size to be $15\times15$ for a lower model complexity.

  \begin{figure*}[!tp]
  \centering
  \includegraphics[width=1.00\textwidth]{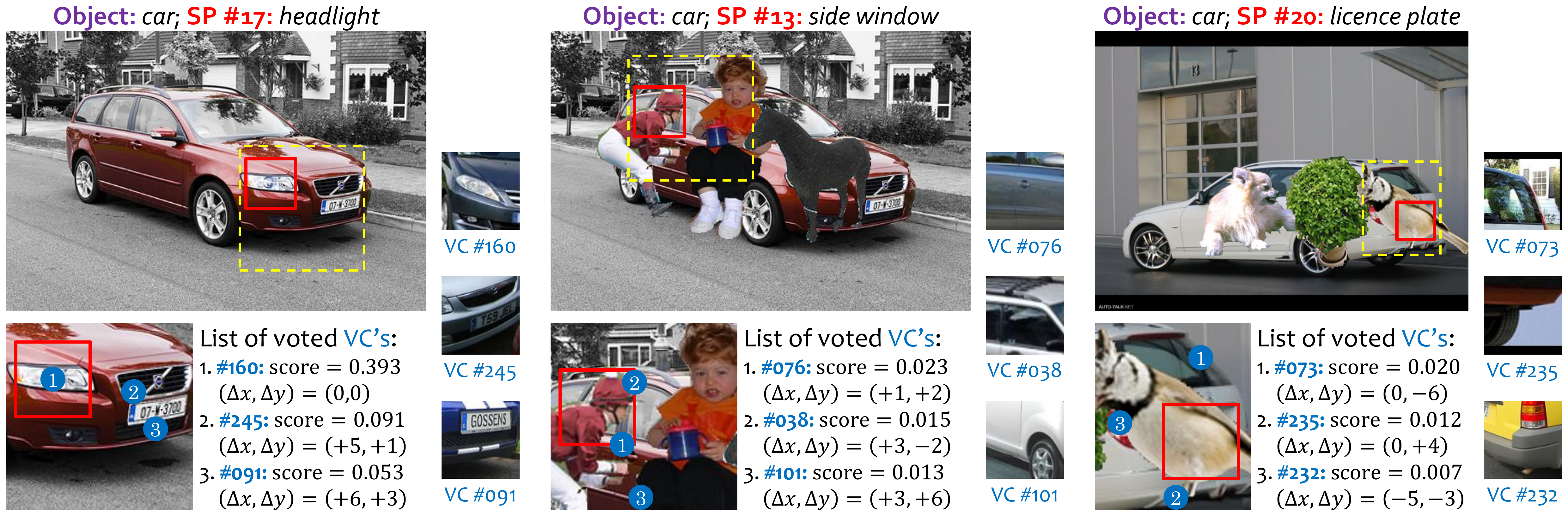}
  \caption{
  DeepVoting allows us to explain the detection results. In the example of heavy occlusion (the third column), the target semantic part, {\em i.e.}, the {\em licence plate} on a {\em car}, is fully occluded by a {\em bird}. With the help of some visual concepts (blue dots), especially the $73$-rd VC (also displayed in Figure~\ref{Fig:Visualization}), we can infer the position of the occluded semantic part (marked in red). Note that we only plot the $3$ VC's with the highest scores, regardless the number of voting VC's can be much larger.
  }
  \label{Fig:Explanation}
  \end{figure*}

  To verify our motivation that Faster-RCNN suffers downgraded performance in both the proposal network and the classifier, we investigate both the recall of the proposals and the accuracy of the classifier. Results are summarized in Table~\ref{Tab:Diagnosis}. First, we can see that the recall of the proposals goes down significantly as the occlusion level goes up, since the objectness of the semantic part region may become weaker due to the randomly placed occluders. Thus the second stage, {\em i.e.}, classification, has to start with a relatively low-quality set of candidates. In the second part, we add the ground-truth bounding boxes to the existing proposals so that the recall is $100\%$, feed these candidates to the classifier, and evaluate its performance on the occluded images. Even with such benefits, Faster-RCNN still produces unsatisfying detection accuracy. For example, in detecting the semantic parts of a {\em bicycle} at the highest occlusion level ({\bf L3}), making use of the additional proposals from ground-truth bounding boxes merely improves the detection accuracy from $41.1\%$ to $44.7\%$, which is still much lower than the number $62.5\%$ produced by DeepVoting+. This implies that the classifier may be confused since the occluder changes the appearance of the proposals.

  \subsection{Visualizing Visual Concepts and Heatmaps}
  \label{Experiments:NonOcclusion:Visualization}

  In Figure~\ref{Fig:Visualization}, we show some typical examples of the learned visual concepts and spatial heatmaps. The visualization of visual concepts follows the approach used in~\cite{Wang_2017_VC_journal}, which finds $10$ most significant responses on each convolutional filter, {\em i.e.}, the matching template, traces back to the original image lattice, and crops the region corresponding to the neuron at the {\em pool-4} layer. To show different spatial heatmaps, we randomly choose some relevant pairs of visual concept and semantic part, and plot the convolutional weights of the voting layer for comparison. We see that the learned visual concepts and spatial heatmaps are semantically meaningful, even though there is only semantic part level supervision during training.

  \subsection{Explaining the Detection Results}
  \label{Experiments:Explanation}
  Finally, we show an intriguing benefit of our approach, which allows us to explain the detection results. In Figure~\ref{Fig:Explanation}, we display three examples, in which the target semantic parts are not occluded, partially occluded and fully occluded, respectively. DeepVoting can infer the occluded semantic parts, and is also capable of looking up the voting (supporting) visual concepts for diagnosis, to dig into errors and understand the working mechanism of our approach.

  \section{Conclusions}
  \label{Conclusions}
  In this paper, we propose a robust and explainable deep network, named DeepVoting, for semantic part detection under partial occlusion. The intermediate visual representations, named {\em visual concepts}, are extracted and used to vote for semantic parts via two convolutional layers. The spatial relationship between visual concepts and semantic parts is learned from a occlusion-free dataset and then transferred to the occluded testing images. DeepVoting is evaluated on both the VehicleSemanticPart dataset and the  VehicleOcclusion dataset, and shows comparable performance to Faster-RCNN in the non-occlusion scenario, and superior performance in the occlusion scenario. If context is utilized, {\em i.e.}, DeepVoting+, this framework outperforms both DeepVoting and Faster-RCNN significantly under all scenarios. Moreover, our approach enjoys the advantage of being explainable, which allows us to diagnose the semantic parts detection results by checking the contribution of each voting visual concepts.

  In the future, we plan to extend DeepVoting to detect semantic parts of non-rigid and articulated objects like animals. Also, we plan to perform object-level detection under occlusion by combining these semantic cues.

  \section*{Acknowledgements}
  \label{Acknowledgements}

  This research was supported by ONR grant N00014-15-1-2356, and also by the Center for Brains, Minds, and Machines (CBMM),
  funded by NSF STC award CCF-1231216. We thank Chenxi Liu, Zhuotun Zhu, Jun Zhu, Siyuan Qiao, Yuyin Zhou and Wei Shen for instructive discussions.

  {\small
  \bibliographystyle{ieee}
  \bibliography{refs}
  }
  \end{document}